\documentclass[runningheads]{llncs}
\usepackage[T1]{fontenc}
\usepackage{graphicx,verbatim}
\usepackage{pifont}
\usepackage{booktabs}
\usepackage{multirow}
\usepackage{makecell}
\usepackage{amsmath}
\usepackage{amssymb}
\usepackage{marvosym}
\usepackage{color}

\usepackage[colorlinks=true,linkcolor=black,citecolor=black,urlcolor=black]{hyperref}

\begin{document}

\title{OPGAgent: An Agent for Auditable Dental Panoramic X-ray Interpretation}

\author{Zhaolin Yu\inst{1,2} \and
Litao Yang\inst{1,2}(\textrm{\Letter}) \and
Ben Babicka\inst{3} \and
Ming Hu\inst{1,2} \and
Jing Hao\inst{4} \and
Anthony Huang\inst{5} \and
James Huang\inst{5} \and
Yueming Jin\inst{6} \and
Jiasong Wu\inst{7} \and
Zongyuan Ge\inst{1,2,5}(\textrm{\Letter})}
\authorrunning{Z. Yu et al.}
\institute{AIM for Health Lab, Faculty of Information Technology, Monash University, Melbourne, Australia \and
Faculty of Information Technology, Monash University, Melbourne, Australia \and
Curae Health, Melbourne, Australia \and
Faculty of Dentistry, The University of Hong Kong, Hong Kong \and
Airdoc-Monash Research, Monash University, VIC 3800, Australia \and
National University of Singapore, Singapore City, Singapore \and
Southeast University, Nanjing, China \\
\email{litao.yang@monash.edu, zongyuan.ge@monash.edu}}

\maketitle

\begin{abstract}
Orthopantomograms (OPGs) are the standard panoramic radiograph in dentistry, used for full-arch screening across multiple diagnostic tasks. While Vision Language Models (VLMs) now allow multi-task OPG analysis through natural language, they underperform task-specific models on most individual tasks. Agentic systems that orchestrate specialized tools offer a path to both versatility and accuracy, this approach remains unexplored in the field of dental imaging. To address this gap, we propose OPGAgent, a multi-tool agentic system for auditable OPG interpretation. OPGAgent coordinates specialized perception modules with a consensus mechanism through three components: (1) a Hierarchical Evidence Gathering module that decomposes OPG analysis into global, quadrant, and tooth-level phases with dynamically invoking tools, (2) a Specialized Toolbox encapsulating spatial, detection, utility, and expert zoos, and (3) a Consensus Subagent that resolves conflicts through anatomical constraints. We further propose OPG-Bench, a structured-report protocol based on (Location, Field, Value) triples derived from real clinical reports, which enables a comprehensive review of findings and hallucinations, extending beyond the limitations of VQA indicators. On our OPG-Bench and the public MMOral-OPG benchmark, OPGAgent outperforms current dental VLMs and medical agent frameworks across both structured-report and VQA evaluation. Code will be released upon acceptance.

\end{abstract}

\section{Introduction}

The Orthopantomogram (OPG) captures the maxilla, mandible, and full dentition in a single panoramic exposure and is routinely used for screening, diagnosis, and treatment planning~\cite{dental_panoramic_survey}. Deep learning models now exist for individual steps of OPG analysis, including caries detection~\cite{bayraktar_caries}, alveolar bone loss assessment~\cite{kurt_boneloss}, tooth segmentation~\cite{tooth_assessment,sts_miccai}, and tooth enumeration~\cite{dentex}. However, as recent reviews note~\cite{dental_automation_challenges,dental_dl_decade}, since these separate models cannot connect with each other, clinicians must use multiple tools one by one, making their workflow inefficient. This motivates the development of a unified dental framework capable of concurrently addressing diverse diagnostic tasks.

To address this need for multi-tasking, Vision Language Models (VLMs) take a different approach by framing diverse visual tasks as a unified text generation problem. Dental VLMs such as DentalGPT~\cite{dentalgpt} and OralGPT-Omni~\cite{oralgpt}, together with general medical VLMs like LLaVA-Med~\cite{llavamed} and MedGemma~\cite{medgemma}, can perform pathology identification, anatomical description, and report generation via multimodal information. Their accuracy on individual tasks, however, lags behind that of specialized models, where local spatial cues dominate~\cite{cnn_vs_transformer}. Therefore, overcoming this performance limitation while sustaining multi-task adaptability remains a challenge.

AI agents~\cite{react} bridge this gap by coordinating specialized models and external tools, integrating high-precision multi-modal outputs into a unified framework. MedAgents~\cite{medagents} and MDAgents~\cite{mdagents} use collaborative reasoning for clinical decision-making; MedAgent-Pro~\cite{medagentpro} adds external perception tools to improve diagnostic accuracy. While these frameworks excel in specific modalities or broad clinical applications, they are not designed to capture the specialized nuances of dentistry. Effective processing of dental imaging relies heavily on domain knowledge like FDI notation~\cite{dentex,fdi_notation} and specialized mechanisms such as dynamic Region of Interest (ROI) cropping. Our experiments reveal a significant performance gap when applying medical agent designs (Table~\ref{tab:results}) to specialized dental tasks, underscoring the necessity for a domain-specific agent.

To reliably test these models, we need a complete benchmark. However, existing VQA benchmarks, whether closed-ended or open-ended, measure precision only on the questions asked; recall depends entirely on whether those questions cover all clinically relevant findings. As a result, (1) any finding type absent from the question set is invisible to evaluation, and (2) hallucination severity cannot be quantified, since fabricated findings in unprompted regions go undetected~\cite{med_hallucination,marine}. Furthermore, this question-and-answer paradigm fundamentally diverges from real-world clinical practice. Dentists do not analyze an OPG by answering isolated prompts; rather, they conduct comprehensive visual screenings to generate structured, holistic reports. Consequently, current benchmarks fail to reflect a model's true diagnostic utility in an actual clinical workflow.

To address these issues, we propose OPGAgent, a multi-tool agentic system for auditable OPG interpretation that combines specialized perception modules with a consensus mechanism. The system has three components: (1)~a Hierarchical Evidence Gathering module that decomposes analysis into global, quadrant, and tooth-level phases; (2)~a Specialized Toolbox encapsulating four categories; and (3)~a Consensus Subagent that aggregates votes from detection tools and expert zoos and resolves conflicts through anatomical constraints. Furthermore, to align our evaluation with real clinical workflows rather than isolated prompts, we further propose OPG-Bench, a structured-report protocol utilizing (Location, Field, Value) triples to overcome the failure of VQA in reflecting true diagnostic capabilities. In summary, our contributions are threefold: (1)~\textbf{OPGAgent}, the first agentic system specifically designed for OPG interpretation, featuring Hierarchical Evidence Gathering, a Specialized Toolbox, and a Consensus Subagent; (2)~\textbf{OPG-Bench}, a novel evaluation protocol derived from real clinical reports that explicitly audits both pathological findings and hallucinations; and (3)~\textbf{State-of-the-art performance}, demonstrating that OPGAgent outperforms all baselines on both OPG-Bench and MMOral-OPG~\cite{mmoral}, achieving higher F1 scores with fewer false positives.

\section{OPGAgent}

\subsection{Overview.} 
\begin{figure}[t]
\centering
\includegraphics[width=\textwidth]{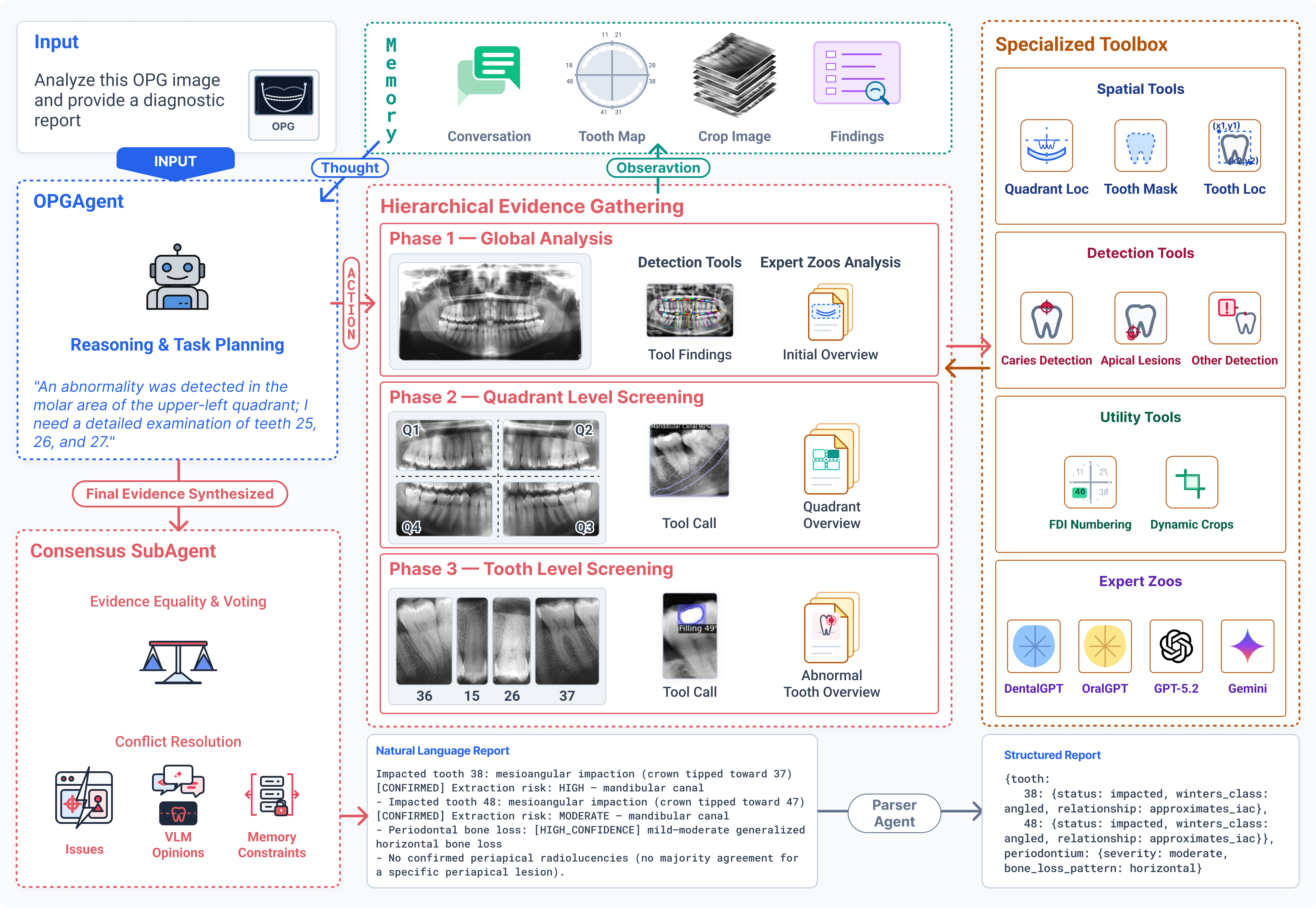}
\caption{Overview of OPGAgent. The Agent orchestrates three modules: Hierarchical Evidence Gathering, Specialized Toolbox, and Consensus Subagent.}
\label{fig:framework}
\end{figure}

Fig.~\ref{fig:framework} illustrates the workflow of OPGAgent. Given an input OPG image $I$, OPGAgent is a planner powered by GPT-5.2 following the ReAct paradigm~\cite{react} that orchestrates three modules: Hierarchical Evidence Gathering, Specialized Toolbox, and Consensus Subagent. In the Hierarchical Evidence Gathering module, OPGAgent decomposes the analysis into three phases: global dentition, quadrant-level screening, and tooth-level screening, with findings stored in Memory. The Specialized Toolbox wraps perception models and VLM experts as agent-callable tools in four categories: spatial, detection, utility, and expert zoos. The Consensus Subagent aggregates evidence from detection tools and expert zoos, resolves conflicts using anatomical constraints, and commits only well-supported findings to the final report. 

\subsection{Hierarchical Evidence Gathering}

To generate refined, verifiable findings, we design a Hierarchical Evidence Gathering module that progressively refines results through three phases, from full images to the tooth level, and stores all findings in Memory.

\noindent\textbf{Phase 1: Global Analysis.} First, the agent queries VLM experts on the full image to generate an initial reading. Simultaneously, it invokes detection tools to establish the anatomical baseline: total tooth count, missing teeth, and FDI notation mapping~\cite{fdi_notation} for each tooth. This phase creates a structured coordinate system in Memory for all subsequent phases.

\noindent\textbf{Phase 2: Quadrant-Level Screening.} Using the global context in Memory, OPGAgent scans each quadrant (Q1--Q4). It uses coordinates from Phase~1 to generate dynamic quadrant crops, which are sent to VLM experts to screen for gross pathologies such as alveolar bone loss or large lesions. The marked regions will be sent to Phase~3 for tooth-level investigation.

\noindent\textbf{Phase 3: Tooth-Level Screening.} For detailed pathologies (e.g., caries, impaction status), OPGAgent crops dynamic ROIs using coordinates in Memory. Because bounding boxes preserve local context, relatively higher resolution crops are sent to VLM experts for assessment. If Phase~2 flags a potential issue in a region where Phase~1 detected no tooth (e.g., a root remnant), OPGAgent requests a quadrant crop from independent anatomical detection (not tooth coordinates), allowing VLMs to recover false negatives missed by the tooth-level scan.

\subsection{Specialized Toolbox}

To provide structured visual perception for OPGAgent, we implement perception modules and expose them as agent-friendly tools. These tools are powered by four underlying model types: (1) YOLO~\cite{yolo} trained on public dental datasets~\cite{dentex} for quadrant and tooth detection; (2) open-source dental vision experts from OralGPT-Omni~\cite{oralgpt}, based on MaskDINO~\cite{maskdino} and DINO~\cite{dino_det}, for maxillary sinus, mandibular canal, and pathology detection; (3) MedSAM~\cite{medsam} for tooth mask segmentation; and (4) multiple VLM for whole-image or ROI-level analysis.

\textbf{Spatial Tools.} These tools return masks or bounding boxes with coordinates for teeth, quadrants, and detected findings. These coordinates form the spatial reference for all subsequent phases.

\textbf{Detection Tools.} These tools detect specific or all abnormal conditions using various pathological models (caries, periapical lesions, apical fillings, etc.) and determine which teeth are associated with these abnormalities.

\textbf{Utility Tools.} This suite manages FDI numbering, ROI extraction, and spatial reasoning. To assess clinical risks such as the proximity of a tooth root to the mandibular canal, the agent computes the minimum contour distance $d(m_i, m_j)$ between two polygon masks $m_i$ and $m_j$:
\begin{equation}
d(m_i, m_j) = \min_{p \in \partial m_i,\, q \in \partial m_j} \|p - q\|_2
\end{equation}
where $d(m_i, m_j)=0$ if the polygons intersect. For matching diseases to teeth, the agent prioritizes Intersection over Union (IoU), using the Euclidean distance between box centers as a fallback when IoU is below a threshold.

\textbf{Expert Zoos.} These tools call DentalGPT, OralGPT-Omni, GPT-5.2, and Gemini-3-Flash to collect multiple expert opinions on whole images, whole images with bounding box, or specific ROIs. These opinions form the evidence base for the Consensus Subagent.

\subsection{Consensus Subagent}

To mitigate hallucination risks in generative models, we design a Consensus Subagent that aggregates evidence from detection tools and expert zoos.

\noindent\textbf{Evidence Aggregation.} The Consensus Subagent collects opinions from $N$ sources and confirms a finding when $\geq 3$ sources agree or $\geq 2$ sources report the same finding. All sources receive equal voting weight so that rare conditions missed by rule-based detectors can still pass through VLM votes.

\noindent\textbf{Conflict Resolution.} Disagreements often concern exact attributes rather than the presence of a finding. When a finding is confirmed by majority vote but sources disagree on its attributes (e.g., tooth number or severity grade), the subagent checks the conflicting claims against hard constraints provided by the detection tools. For instance, if VLMs agree on a ``lower left implant'' but assign different tooth numbers, the subagent consults the detection tool's FDI coordinate map to assign the finding to the spatially correct tooth, correcting the VLM output with deterministic coordinates.

\subsection{Structured Reporting and Protocol Design}

\subsubsection{Hierarchical Clinical Ontology.} To align the agent's output with real-world clinical triage, we formalize the OPG interpretation into a structured, anatomy centric schema. We define a comprehensive dental report as a set of pathological findings $\mathcal{S} = \{t_1, t_2, \dots, t_N\}$, where each finding is abstracted as an information triple $t_i = (l_i, f_i, v_i) \in \mathcal{L} \times \mathcal{F} \times \mathcal{V}$. Here, $\mathcal{L}$ denotes the anatomical location space strictly adhering to the FDI World Dental Federation notation (ISO 3950) for teeth, alongside global regions (e.g., TMJ, maxillary sinus). $\mathcal{F}$ represents the clinical field, and $\mathcal{V}$ is the categorical value space. To ensure clinical completeness, $\mathcal{V}$ is constructed upon standardized dental guidelines, mapping complex numerical scales into semantic severity levels (e.g., ICDAS~\cite{icdas} for caries, AAP/EFP 2017~\cite{tonetti_perio} for periodontitis, and PAI~\cite{pai_orstavik} for periapical health). Following the sparse representation principle of clinical reporting, $\mathcal{S}$ only encapsulates anomalous findings; normal anatomical structures are intentionally omitted.

\noindent\textbf{Constrained Parser Agent.} A dedicated Parser Agent transforms the system's Memory output into the formal $\mathcal{S}$ space. To eliminate hallucinations and enforce schema adherence, it utilizes constrained decoding—via JSON mode, Pydantic validation, and few-shot learning. This deterministically maps findings into valid $(l_i, f_i, v_i)$ triples, strictly preventing unsupported free-text diagnoses.

\noindent\textbf{Triple-based Hierarchical Evaluation (OPG-Bench).} Overcoming VQA limitations, we evaluate findings as $(location, field, value)$ triples via two complementary metrics. The \textbf{Exact Match} requires perfect alignment of all elements. To reflect clinical workflows, we also compute a step-wise partial match: \textbf{Detection} of pathology presence (Precision/Recall/F1), \textbf{Localization} of the anatomical target $l_i$ (Precision/Recall/F1), and \textbf{Classification} of the grade/type $v_i$ (Accuracy). Each step conditions on the previous one. This hierarchical protocol cleanly disentangles visual perception from clinical reasoning.

\section{Experiments}

\textbf{Dataset.} The OPG-Bench dataset comprises 1{,}009 ano\-nymized OPGs with paired clinical and structured reports, and 5,219 unguided VQA pairs; evaluation follows the triple-based protocol defined in \S2.4. Sourced from multiple clinics, the dataset is restricted to patients $\geq 16$ years to focus on permanent dentition. Ground truths derive from real clinical reports and are validated via manual spot-checks. Since real-world reports typically prioritize chief complaints over incidental findings, our reported False Positive rates may be slightly inflated when the agent detects undocumented yet valid pathologies.

\noindent\textbf{Evaluation Protocol.} We evaluate OPGAgent against state-of-the-art general VLMs, dental-specific VLMs, and medical agents. To isolate our framework's architectural contribution, MedAgent-Pro \cite{medagentpro} is configured with the same LLM and tools as OPGAgent. To decouple diagnostic accuracy from formatting, we employ a unified generation-evaluation pipeline. All models receive identical prompts (temperature 0.3) to generate natural-language reports. A shared Parser Agent then converts these into structured formats---ensuring parser bias applies equally across methods---followed by manual filtering to retain only valid clinical findings.

\begin{table}[ht]
\centering
\caption{Quantitative comparison on OPG-Bench dataset.}
\label{tab:results}
\footnotesize
\resizebox{\textwidth}{!}{%
\setlength{\tabcolsep}{2pt}
\begin{tabular}{lcccccccccccc}
\toprule
\multirow{2}{*}{Methods} &
\multicolumn{5}{c}{E.M.} &
\multicolumn{3}{c}{Det.} &
\multicolumn{3}{c}{Loc.} &
\multicolumn{1}{c}{Cls.} \\
\cmidrule(lr){2-6}\cmidrule(lr){7-9}\cmidrule(lr){10-12}\cmidrule(lr){13-13}
& Score & P & R & F1 & FP$\downarrow$ & P & R & F1 & P & R & F1 & Acc \\
\midrule
Gemini-3-Flash & \underline{0.428} & 0.276 & \textbf{0.451} & \underline{0.343} & 10.58 & 0.440 & \textbf{0.596} & \underline{0.506} & 0.311 & \textbf{0.488} & 0.380 & \underline{0.796} \\
Kimi-k2.5 & 0.399 & 0.252 & 0.345 & 0.291 & 9.20 & 0.481 & \underline{0.500} & 0.490 & 0.348 & \underline{0.463} & \underline{0.397} & 0.764 \\
Gemini-3-Pro & 0.399 & \underline{0.323} & 0.294 & 0.308 & 5.60 & 0.556 & 0.431 & 0.485 & \underline{0.458} & 0.319 & 0.376 & 0.732 \\
GPT-5.2 & 0.357 & 0.283 & 0.273 & 0.278 & 6.17 & \underline{0.624} & 0.359 & 0.456 & 0.392 & 0.193 & 0.258 & 0.754 \\
Qwen3-VL-235B & 0.230 & 0.106 & 0.191 & 0.137 & 14.50 & 0.316 & 0.436 & 0.366 & 0.107 & 0.175 & 0.133 & 0.614 \\
Qwen3-VL-8B & 0.197 & 0.102 & 0.070 & 0.083 & 5.50 & 0.409 & 0.259 & 0.317 & 0.175 & 0.124 & 0.145 & 0.635 \\
\hline
DentalGPT & 0.295 & 0.128 & 0.200 & 0.156 & 12.22 & 0.464 & 0.483 & 0.473 & 0.241 & 0.273 & 0.256 & 0.715 \\
OralGPT-Omni & 0.157 & 0.094 & 0.046 & 0.062 & \textbf{4.02} & 0.457 & 0.151 & 0.227 & 0.145 & 0.075 & 0.099 & 0.603 \\
\hline
MedAgent-Pro & 0.278 & 0.171 & 0.183 & 0.177 & 8.00 & 0.412 & 0.438 & 0.425 & 0.206 & 0.204 & 0.205 & 0.632 \\
\hline
OPGAgent & \textbf{0.497} & \textbf{0.431} & \underline{0.415} & \textbf{0.423} & \underline{4.89} & \textbf{0.715} & 0.456 & \textbf{0.557} & \textbf{0.625} & 0.376 & \textbf{0.469} & \textbf{0.801} \\
\bottomrule
\end{tabular}%
}
\footnotesize
\noindent \textit{Score $= 0.5\times$EM\_F1 $+ 0.2\times$Det\_F1 $+ 0.2\times$Loc\_F1 $+ 0.1\times$Cls\_Acc (balancing exact and partial matches). E.M.\ (Exact Match), Det.\ (Presence), Loc.\ (Location), Cls.\ (Grade/Type), FP (False Positives/Case).}
\end{table}

\noindent\textbf{Results on OPG-Bench.} We evaluate OPGAgent against general VLMs, and dental VLMs~\cite{dentalgpt,oralgpt} on the OPG-Bench dataset. As shown in Table~\ref{tab:results}, OPGAgent reaches an exact-match F1 of 42.3\% and an aggregate score of 49.7\%, surpassing Gemini-3-Flash and DentalGPT. While Gemini-3-Flash obtains higher recall (45.1\%), its precision drops to 27.6\% with 10.58 false positives per case; OPGAgent maintains precision at 43.1\% while keeping false positives at 4.89. Domain-specific models like OralGPT-Omni minimize false positives (4.02) but at the cost of low coverage (F1 6.2\%). OPGAgent thus balances sensitivity and reliability better than single-pass approaches.

\begin{table}[t]
\centering
\caption{Performance comparison on VQA benchmarks.}
\label{tab:vqa_combined}
\footnotesize
\resizebox{\textwidth}{!}{%
\setlength{\tabcolsep}{2pt}
\begin{tabular}{l|c|cccc|c|ccccc}
\toprule
\multirow{2}{*}{Methods} & \multicolumn{5}{c|}{\textbf{MMOral-OPG}} & \multicolumn{6}{c}{\textbf{OPG-Bench (VQA)}} \\
& Acc. & Tee. & Pat. & Jaw & HisT & Acc. & Mis. & Muc. & Bone & Anat. & Api. \\
\midrule
Gemini-3-Flash & 48.47\% & 54.1\% & 42.6\% & 59.7\% & 66.2\% & 59.9\% & 48.4\% & 47.9\% & 60.0\% & 65.1\% & \textbf{42.2\%} \\
GPT-5.2 & 43.58\% & 37.2\% & 32.4\% & 65.9\% & 52.1\% & 51.5\% & 48.8\% & 44.7\% & 50.5\% & 65.1\% & 27.7\% \\
OralGPT-Omni & 59.06\% & 53.5\% & 44.6\% & \textbf{77.5\%} & 69.7\% & 36.1\% & 13.6\% & 15.7\% & 23.2\% & 30.8\% & 17.0\% \\
DentalGPT & 48.27\% & 42.3\% & 40.5\% & 65.1\% & 47.2\% & 27.8\% & 11.4\% & 22.2\% & 4.2\% & 13.0\% & 17.5\% \\
\hline
\textbf{OPGAgent} & \textbf{62.53\%} & \textbf{59.7\%} & \textbf{52.7\%} & 72.1\% & \textbf{71.8\%} & \textbf{61.2\%} & \textbf{67.0\%} & \textbf{65.2\%} & \textbf{76.8\%} & \textbf{80.8\%} & 35.0\% \\
\bottomrule
\end{tabular}%
}
\footnotesize
\noindent \textit{Tee. (Teeth-related), Pat. (Pathology), HisT (History \& Treatment). Mis. (Missing Teeth), Muc. (Mucosal Change), Bone (Bone Variant), Anat. (Anatomical Variants), Api. (Apical Status).}
\end{table}

\noindent\textbf{Results on VQA Benchmarks.} As shown in Table~\ref{tab:vqa_combined}, OPGAgent leads on both benchmarks (62.53\% on MMOral-OPG, 61.2\% on OPG-Bench). On MMOral-OPG, OralGPT-Omni scores highest on Jaw questions (77.5\%) but falls behind on Teeth and Pathology categories, where OPGAgent's multi-tool pipeline provides clearer gains. On OPG-Bench, domain-specific VLMs (OralGPT-Omni 36.1\%, DentalGPT 27.8\%) score much lower than general VLMs, despite OralGPT-Omni performing well on its own MMOral-OPG benchmark (59.06\%). This gap suggests that models trained on generated QA pairs may not fully capture the distribution of real clinical reports.

\begin{table}[t]
\centering
\caption{Ablation study on OPG-Bench with cumulative component additions.}
\label{tab:ablation}
\footnotesize
\setlength{\tabcolsep}{4pt}
\begin{tabular}{ccc cccc}
\toprule
\multicolumn{3}{c}{\textbf{Components}} & \multicolumn{4}{c}{\textbf{Metrics}} \\
\cmidrule(lr){1-3} \cmidrule(lr){4-7}
\makecell{\textbf{Expert}\\\textbf{Zoos}} & \makecell{\textbf{Spatial}\\\textbf{Tools}} & \makecell{\textbf{Detection}\\\textbf{Tools}} & \textbf{Precision} & \textbf{Recall} & \textbf{F1} & \textbf{FP/Case}$\downarrow$ \\
\midrule
\ding{55} & \ding{55} & \ding{55} & 28.32\% & 27.25\% & 27.78\% & 6.17 \\
\ding{51} & \ding{55} & \ding{55} & 38.62\% & 16.64\% & 23.25\% & 2.37 \\
\ding{51} & \ding{51} & \ding{55} & 37.13\% & 35.98\% & 36.55\% & 5.47 \\
\midrule
\ding{51} & \ding{51} & \ding{51} & \textbf{43.15\%} & \textbf{41.48\%} & \textbf{42.30\%} & \textbf{4.89} \\
\bottomrule
\end{tabular}\\[4pt]
\parbox{0.85\textwidth}{\footnotesize\textit{Expert Zoos (GPT-5.2, Gemini-3-Flash, DentalGPT, OralGPT-Omni), Detection Tools (Pathology Detection), Spatial Tools (Tooth/Quadrant Localization).}}
\end{table}

\noindent\textbf{Ablation Studies.} Table~\ref{tab:ablation} adds components cumulatively. By design, the agent LLM serves only as a planner. Without external tools (Row~1), it is forced to act as the sole diagnostician in a ReAct loop, yielding a 27.78\% F1. Adding \textbf{Expert Zoos} (Row~2) improves precision (38.62\%) and minimizes false positives (6.17 $\rightarrow$ 2.37), but plummets recall (16.64\%) because findings lack exact FDI coordinates and fail strict triple matching. \textbf{Spatial Tools} (Row~3) resolve this grounding bottleneck, restoring recall (35.98\%) and F1 (36.55\%). Finally, \textbf{Detection Tools} (Row~4) provide complementary localized evidence, achieving the peak 42.30\% F1.

\begin{figure}[!htb]
\centering
\includegraphics[width=\textwidth]{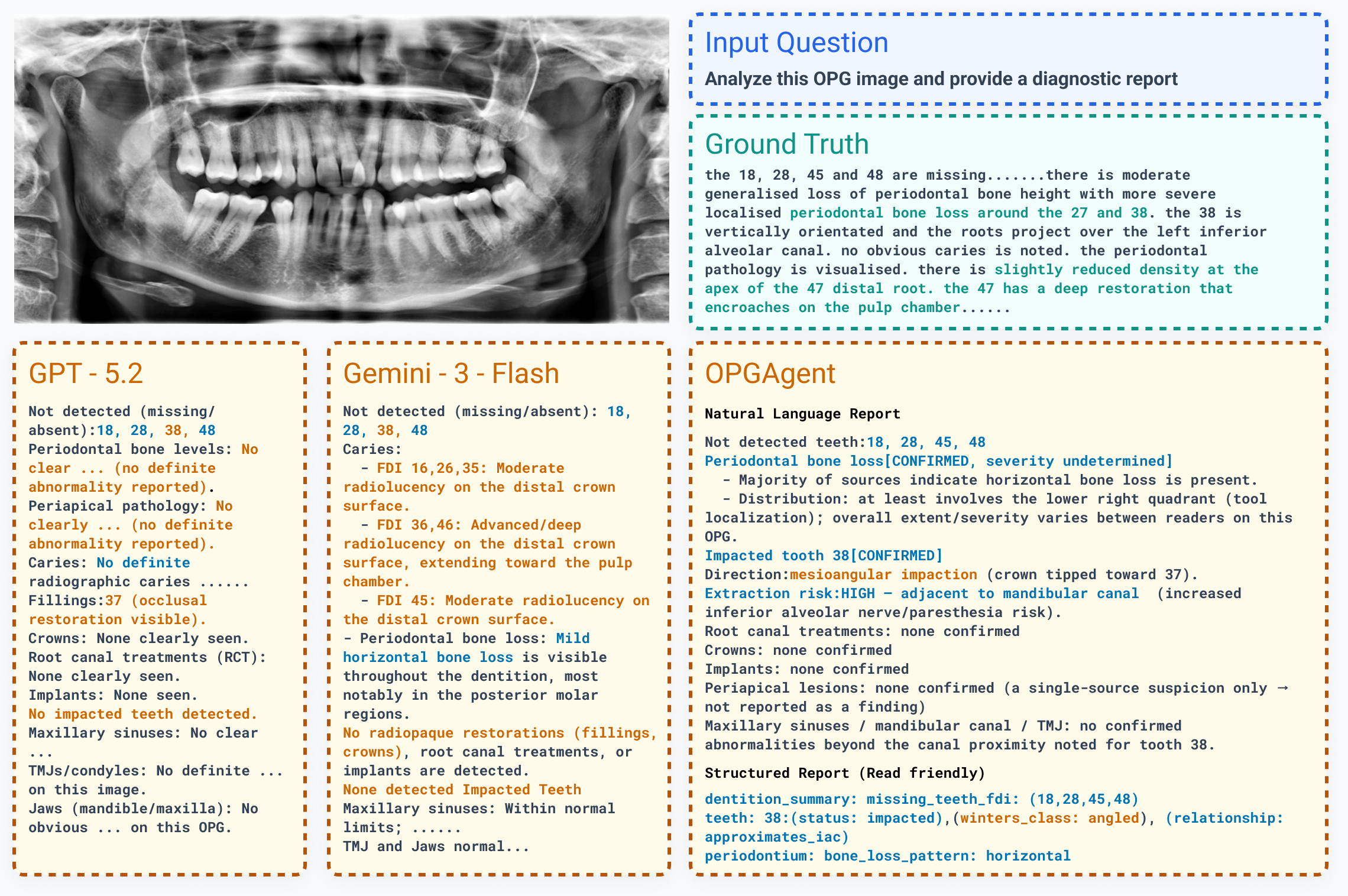}
\caption{Vision comparison. OPGAgent correctly identifies missing teeth, bone loss, and impacted tooth~38 via consensus, while GPT-5.2 and Gemini-3-Flash each miss or hallucinate findings.\textcolor[rgb]{0.0,0.45,0.70}{Blue} for correct; \textcolor[rgb]{0.8,0.4,0.0}{Orange} for wrong; \textcolor[rgb]{0.05, 0.58, 0.53}{Green} for all undetected.} 
\label{fig:qualitative_case}
\end{figure}

\section{Conclusion}

We present OPGAgent, a multi-tool agentic framework for auditable OPG interpretation that combines Hierarchical Evidence Gathering, a Specialized Toolbox, and a Consensus Subagent. Through phased analysis, multi-source voting, anatomical conflict resolution, and structured triple-based evaluation, OPGAgent produces more accurate and auditable dental reports than existing VLMs and agent frameworks. Ablation studies confirm that each module contributes to the overall performance.

\end{document}